\documentclass[10pt,twocolumn,letterpaper]{article}

\usepackage{cvpr}
\usepackage{times}
\usepackage{epsfig}
\usepackage{graphicx}
\usepackage{amsmath}
\usepackage{amssymb}

\usepackage{tabularx}
\usepackage{tablefootnote}
\usepackage{url}
\usepackage{subfigure}
\usepackage{placeins} 
\usepackage{color}
\usepackage[dvipsnames]{xcolor}
\usepackage{multirow}

\newcolumntype{C}[1]{>{\centering\arraybackslash}p{#1}}
\newcommand{\uline}{\underline}

\usepackage[pagebackref=true,breaklinks=true,letterpaper=true,colorlinks,bookmarks=false]{hyperref}

\cvprfinalcopy 

\ifcvprfinal\fi

\begin{document}

\title{CNN-based Patch Matching for Optical Flow with Thresholded Hinge Embedding Loss}

\author{~~Christian Bailer$^1$~~~~~~~~~~~~~~~~~~~~~~Kiran Varanasi$^{1}$~~~~~~~~~~~~~~~~~~~~Didier Stricker$^{1,2}$ \\
{\tt\small Christian.Bailer@dfki.de}~~~~~~~{\tt\small Kiran.Varanasi@dfki.de }~~~~~~~{\tt\small Didier.Stricker@dfki.de}\\
$^1$German Research Center for Artificial Intelligence (DFKI), $^2$University of Kaiserslautern \\
}

\maketitle

\begin{abstract}
Learning based approaches have not yet achieved their full potential in optical flow estimation, where their performance still trails heuristic approaches. 
In this paper, we present a CNN based patch matching approach for optical flow estimation. 
An important contribution of our approach is a novel thresholded loss for Siamese networks. We demonstrate that our loss performs clearly better than existing losses. 
It also allows to speed up training by a factor of 2 in our tests. 
Furthermore, we present a novel way for calculating CNN based features for different image scales, which performs better than existing methods.
We also discuss new ways of evaluating the robustness of trained features for the application of patch matching for optical flow.  
An interesting discovery in our paper is that low-pass filtering of feature maps can increase the robustness of features created by CNNs.
We proved the competitive performance of our approach by submitting it to the KITTI 2012, KITTI 2015 and MPI-Sintel evaluation portals 
where we obtained state-of-the-art results on all three datasets.

\end{abstract}


\section{Introduction}
In recent years, variants of the PatchMatch~\cite{barnes2009patchmatch} approach showed not only to be useful for nearest neighbor field estimation, 
but also for the more challenging problem of large displacement optical flow estimation.
So far, most top performing methods like Deep Matching~\cite{weinzaepfel2013deepflow} or Flow Fields~\cite{bailer2015flow} strongly rely on 
robust multi-scale matching strategies, while they still use engineered features (data terms) like SIFTFlow~\cite{liu2008sift} for the actual matching.

On the other hand, works like~\cite{simo2015discriminative,zagoruyko2015learning} demonstrated the effectiveness of features based on Convolutional Neural Network (CNNs) 
for matching patches.
However, these works did not validate the performance of their features using an actual patch matching approach like \textit{PatchMatch} or \textit{Flow Fields}
that matches all pixels between
image pairs. Instead, they simply treat matching patches as a classification problem between a predefined set of patches.

This ignores many practical issues. For instance, it is important that CNN based features are not only able to distinguish between different patch positions, 
but the position should also be determined accurately. 
Furthermore, the top performing CNN architectures are very slow when used for patch matching as it requires matching several patches for every pixel in the reference image. 
While Siamese networks with $L_2$ distance~\cite{simo2015discriminative} are reasonably fast at testing time and still outperform engineered features
regarding classification, we found that they are usually underperforming engineered features regarding (multi-scale) patch matching. 

We think that this has among other things (see Section~\ref{eval}) 
to do with the convolutional structure of CNNs: as neighboring patches share 
intermediate layer outputs it is much easier for CNNs to learn matches of neighboring patches than non neighboring patches. 
However, due to propagation~\cite{barnes2009patchmatch} correctly matched patches close to each other usually contribute less for patch matching than patches far apart from each other.
 Classification does not differentiate here.

A first solution to succeed in CNN based patch matching is to use pixel-wise batch normalization~\cite{gadot2015patchbatch}. 
While it weakens the unwanted convolutional structure, it is computationally expensive at test time.
Thus, we do not use it. Instead, we improve the CNN features themselves to a level that allows us to outperform existing approaches.

Our first contribution is a novel loss function for the Siamese architecture with $L_2$ distance~\cite{simo2015discriminative}.
We show that the hinge embedding loss~\cite{simo2015discriminative} which is commonly used for Siamese architectures and variants of it 
have an important design flaw: they try to decrease the $L_2$ distance unlimitedly for correct matches, although
very small distances for patches that differ due to effects like illumination changes or partial occlusion are not only very costly 
but also unnecessary, as long as false matches have larger  $L_2$ distances.  
We demonstrate that we can significantly increase the matching quality by relaxing this flaw. 

Furthermore, we present a novel way to calculate CNN based features for the scales of Flow Fields~\cite{bailer2015flow},
which clearly outperforms the original multi-scale feature creation approach, with respect to CNN based features. 
Doing so, an important finding is that low-pass filtering CNN based feature maps robustly improves the matching quality.

Moreover, we introduce a novel matching robustness measure that is tailored for binary decision problems like patch matching 
(while ROC and PR are tailored for classification problems).
By plotting the measure over different displacements and distances between a wrong patch and the correct one  we can reveal interesting properties of different loss functions and scales.
Our main contributions are:

\begin{enumerate}\itemsep0.5pt
 \item
  A novel loss function, that clearly outperforms other state-of-the art losses in our tests and allows to speed up training by a factor of around two.
  \item
  A novel multi-scale feature creation approach tailored for CNN features for optical flow. 
   \item 
  New evaluation measure of matching robustness for optical flow and corresponding plots.
  \item
  We show that low-pass filtering the feature maps created by CNNs improves matching robustness. 
  \item 
  We demonstrate the effectiveness of our approach by obtaining a top performance on all three major evaluation portals
  KITTI 2012~\cite{geiger2013vision}, 2015~\cite{menze2015object} and MPI-Sintel~\cite{butler2012naturalistic}.
  Former learning based approaches always trailed heuristic approaches on at least one of them.  

\end{enumerate}

\section{Related Work}
While regularized optical flow estimation goes back to Horn and Schunck~\cite{horn1981determining}, 
randomized patch matching~\cite{barnes2009patchmatch} is a relatively new field, first successfully applied in approximate 
nearest neighbor estimation where the data term is well-defined.
The success in optical flow estimation (where the data term is not well-defined) started with publications like~\cite{bao2014fast, chen2013large}.
One of the most recent works is Flow Fields~\cite{bailer2015flow}, which showed that with proper multi-scale patch matching, top performing optical flow results
can be achieved.

Regarding patch or descriptor matching with learned data terms, there exists a fair amount of literature~\cite{han2015matchnet, simo2015discriminative, zagoruyko2015learning, simonyan2014learning}. 
These approaches treat matching at an abstract level and do not present a pipeline to solve a problem like optical flow estimation or 3D reconstruction,
although many of them use 3D reconstruction datasets for evaluation. 
Zagoruyko and Komodakis~\cite{zagoruyko2015learning} compared different architectures to compare patches. 
 Simo-Serra et al.~\cite{simo2015discriminative} used the Siamese architecture~\cite{bromley1993signature} with $L_2$ distance. 
 They argued that it is the most useful one for practical applications. 
 
Recently, several successful CNN based approaches for stereo matching appeared~\cite{zbontar2016stereo,luo2016efficient,mayer2015large}.
However, so far there are still few approaches that successfully use learning to compute optical flow.  
Worth mentioning is FlowNet~\cite{fischer2015flownet}. They tried to solve the optical flow problem as a whole 
with CNNs, having the images as CNN input and the optical flow as output.
While the results are good regarding runtime, they are still not state-of-the-art quality.
Also, the network is tailored for a specific image resolution 
and to our knowledge training for large images of several megapixel is still beyond todays computational capacity.

A first approach using patch matching with CNN based features is PatchBatch~\cite{gadot2015patchbatch}.  
They managed to obtain state-of-the-art results on the KITTI dataset~\cite{geiger2013vision}, due to pixel-wise batch normalization and a loss that includes
batch statistics. However, pixel-wise batch normalization is computationally expensive at test time. 
Furthermore, even with pixel-wise normalization their approach trails heuristic approaches on MPI-Sintel~\cite{butler2012naturalistic}.
A recent approach is DeepDiscreteFlow~\cite{Guney2016ACCV} which uses DiscreteFlow~\cite{menze2015discrete} as basis instead of patch matching.
Despite using recently invented dilated convolutions~\cite{luo2016efficient} (we do not use them, yet) they also trail the original DiscreteFlow approach on some datasets.

\renewcommand{\tabcolsep}{1.0pt}
\begin{table}[t]
\footnotesize
 \centering
 \begin{tabular}{|c|c|c|c|c|c|c|c|c|}
  \hline
 Layer  &  1 & 2 & 3 & 4 & 5 & 6 & 7 & 8 \\
 \hline
 Type  & Conv &  MaxPool &  Conv & Conv & MaxPool &Conv & Conv &Conv \\
 \hline
 Input size & 56x56 & 52x52& 26x26& 22x22 & 18x18 & 9x9 & 5x5 & 1x1  \\
 \hline 
 Kernel size & 5x5 &  2x2 &  5x5 & 5x5 & 2x2 & 5x5 & 5x5 & 1x1  \\ 
 \hline
 Out. channels & 64 & 64 & 80 & 160 & 160 & 256 & 512 & 256\\
 \hline
  Stride  & 1  & 2 & 1 & 1 & 2 & 1 & 1 &1     \\ 
 \hline
 Nonlinearity & Tanh   &  - &  Tanh & Tanh & - & Tanh & Tanh & Tanh \\ 
 \hline
 \end{tabular}
 \vspace{0.1cm}
 \caption{The CNN architecture used in our experiments.}
 \label{arch}
\end{table}
\renewcommand{\tabcolsep}{5pt}

\section{Our Approach}

Our approach is based on a Siamese architecture~\cite{bromley1993signature}.
The aim of Siamese networks is to learn to calculate a meaningful feature vector $D(p)$ for each image patch $p$.
During training the $L_2$ distance between feature vectors of matching patches ($p_1 \equiv p^+_2$) is reduced, 
while the $L_2$ distance between feature vectors of non-matching patches ($p_1 \neq p^-_2$) is increased 
(see~\cite{simo2015discriminative} for a more detailed description).

Siamese architectures can be strongly speed up at testing time as neighboring patches in the image share convolutions.
Details on how the speedup works are described in our supplementary material. 
The network that we used for our experiments is shown in Table~\ref{arch}.
Similar to~\cite{brown2011discriminative}, we use Tanh nonlinearity layers as we also have found them to outperform ReLU for Siamese based patch feature creation.

\subsection{Loss Function and Batch Selection}\label{s31}

The most common loss function for Siamese network based feature creation is the  hinge embedding loss:
\begin{equation}
l_h(p_1,p_2) =  \begin{cases} L_2(p_1,p_2), ~~~~~~~~~~~~~~~~~~~~~~~~ p_1 \equiv p_2 \\ max(0,m - L_2(p_1,p_2)),~ p_1 \neq p_2    \end{cases}
\end{equation} 
\begin{equation}
L_2(p_1,p_2) = || D(p_1) - D (p_2) ||_2
\end{equation}

\begin{figure}[t]

  \includegraphics[width=1.0\linewidth]{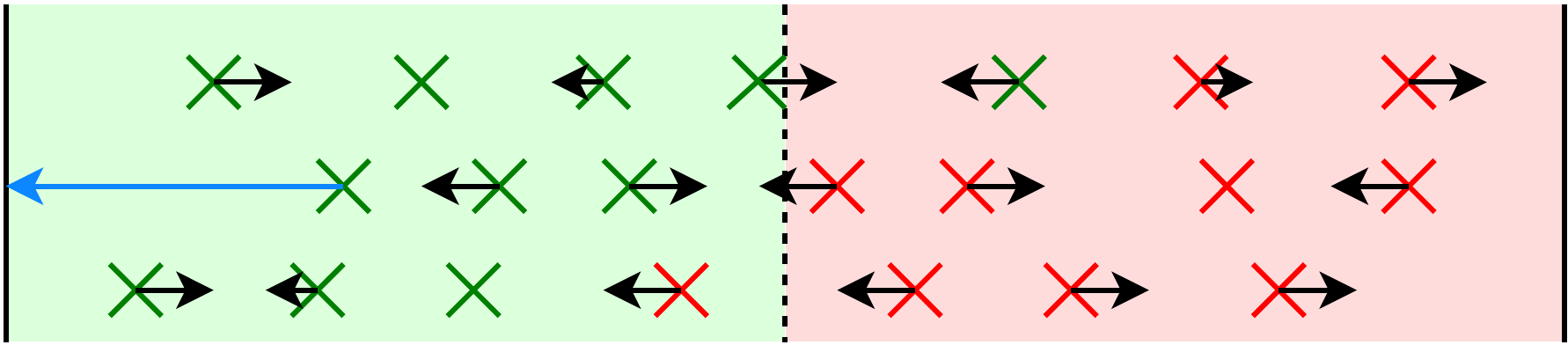}
   \caption{ 
   If a sample is pushed (blue arrow), although it is clearly on the correct side of the decision boundary other samples also move
   due to weight change. If most samples are classified correctly beforehand, this creates more false decision boundary crossings than correct ones.
   $l_h$ performs the unnecessary push, $l_t$ not. 
   }
\label{lossex2}
\end{figure}

It tries to minimize the $L_2$ distance of matching patches and to increase the $L_2$ distance of non-matching patches above $m$.
An  architectural flaw which is not or only poorly treated by existing loss functions is the fact that the loss 
pushes feature distances between matching patches unlimitedly to zero ($ L_2(p_1,p^+_2) \rightarrow 0$). 
We think that training up to very small $L_2$ distances for patches that differ due to effects like rotation or motion blur is very costly -- it 
has to come at the cost of failure for other pairs of patches. A possible explanation for this cost is shown in Figure~\ref{lossex2}.
As a result, we introduce a modified hinge embedding loss with threshold $t$ that stops the network from minimizing $L_2$  distances too much:

\begin{equation}
\label{Eq2}
l_t(p_1,p_2) =  \begin{cases}  max (0, L_2(p_1,p_2)- t), ~~~~~~~~~~~~~~~ p_1 \equiv p_2 \\ max(0,m - L_2(p_1,p_2)- t),~~~~~~ p_1 \neq p_2    \end{cases}
\end{equation} 
We add $t$ also to the second equation to keep the ``virtual decision boundary''  at $m/2$. 
This is not necessary but makes comparison between different $t$ values fairer.

As our goal is a network that creates features with the property $L_2(p_1,p^+_2)<L_2(p_1,p^-_2)$
one might argue that it is better to train this property directly. 
A known function to do this is a gap based loss~\cite{han2015matchnet,wohlhart2015learning}, that only keeps a gap in the $L_2$ distance between matching and non-matching pairs:

\begin{equation}
\begin{split}
\label{Eq3}
l_g(p_1,p^+_2) =   max (0,  L_2(p_1,p^+_2) - L_2(p_1,p^-_2) + g  ), \\
p_1 \equiv p^+_2 \cap  p_1 \neq p^-_2 ~~~~~~~~~~~~~~~~~~~~~~~~~~~~~~~~~~~~~~~~~~~~~~~~~~
\end{split}
\end{equation}
 $l_g(p_1,p^-_2)$ is set to $-l_g(p_1,p^+_2)$ (reverse gradient). 
 While $l_g$ intuitively seems to be better suited for the given problem than $l_t$, we will show in Section~\ref{eval} why this is not the case.
 There we will also compare $l_t$ to further loss functions.

The given loss functions have in common that the loss gradient is sometimes zero. 
Ordinary approaches still back propagate a zero gradient. 
This not only makes the approach slower than necessary, but also leads to a variable effective batch size of training samples, that are
actually back propagated. 
This is a limited issue for the hinge embedding loss $l_h$, where only $\approx 25\%$ of the training samples obtain a zero gradient in our tests.
However, with $l_t$ (and suitable $t$) more than 80\% of the samples obtain a zero gradient.

As a result, we only add training samples with a non-zero loss to a batch. All other samples are rejected without back propagation. 
This not only increases the training speed by a factor of around two in our tests, but also improves the training quality by avoiding
variable effective batch sizes.

\subsection{Training}
Our training set consists of several pairs of images ($I_1$, $I_2 \in I_{all} $) with known optical flow displacement between their pixels.
We first subtract the mean from each image and divide it by its standard derivation.
To create training samples, we randomly extract patches $p_1 \in I_1$ and their corresponding matching patches $p^+_2 \in I_2$, $p_1 \equiv  p^+_2$ for positive training samples.
For each $p_1$, we also extract one non-matching patch $p^-_2 \in I_2$, $p_1 \neq p^-_2$ for negative training samples. 
Negative samples $p^-_2$ are sampled from a distribution $N(p^+_2)$ that prefers patches close to the matching patch $p^+_2$, 
with a minimum distance to it of 2 pixels, but it also allows to sample patches that are far from $p^+_2$. 
The exact distribution can be found in our supplementary material.

We only train with pairs of patches where the center pixel of $p_1$ is not occluded in the matching patch $p^+_2$.
Otherwise, the network would train the occluding object as a positive match. 
However, if the patch center is visible we expect the network to be able to deal with a partial occlusion. 
We use a learning rate between 0.004  and 0.0004 that decreases linearly in exponential space after each batch i.e. 
$learnRate (t) = e^{-x_t} \rightarrow learnRate (t+1) = e^{-(x_t+\epsilon)}$. 

\begin{figure*}[t]
\centering
  \includegraphics[width=0.99\linewidth]{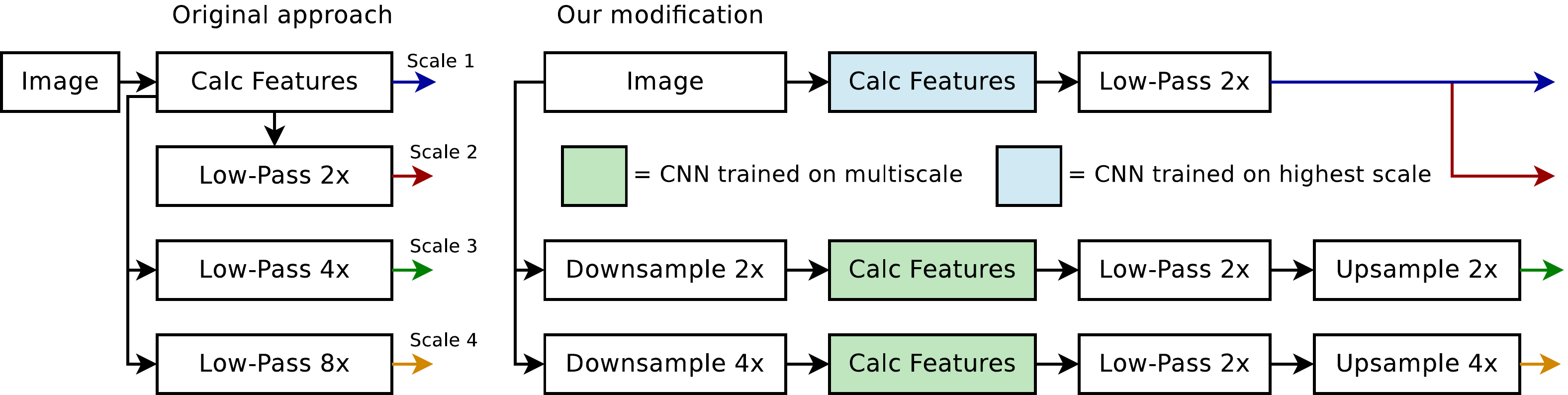}
   \caption{Our modification of feature creation of the \textit{Flow Fields} approach~\cite{bailer2015flow} for clearly better CNN performance.
   Note that \textit{Flow Fields} expects feature maps of all scales in the full image resolution (See~\cite{bailer2015flow} for details).
   Reasons of design decision can be found in Section~\ref{decisions}. }
   \label{ffscales}
\end{figure*}

\subsection{Multi-scale matching}

The \textit{Flow Fields} approach~\cite{bailer2015flow}, which we use as basis for our optical flow pipeline compares patches at 
different scales using scale spaces~\cite{lindeberg1994scale}, i.e. all scales have the full image resolution.
It creates feature maps for different scales by low-pass filtering the feature map of the highest scale (Figure ~\ref{ffscales} left).
For SIFTFlow~\cite{liu2008sift} features used in~\cite{bailer2015flow}, low-pass filtering features (i.e. feature $\rightarrow$ low-pass = feature $\rightarrow$ downsample $\rightarrow$ upsample) 
performs better than recalculating features for each scale on a different resolution (i.e. downsample $\rightarrow$ feature $\rightarrow$ upsample).

We observed the same effect for CNN based features -- even if the CNN is also trained on the lower resolutions. 
However, with our modifications shown in Figure~\ref{ffscales} right (that are further motivated in Section~\ref{eval}), it is possible to obtain better results by recalculating features on different resolutions.
We use a CNN trained and applied only on the highest image resolution for the highest and second highest scale. 
Furthermore, we use a CNN trained on 3 resolutions (100\%, 50\% and 25\%) to calculate the feature maps for the third and fourth scale
applied at 50\% and 25\% resolution, respectively. 
For the multi-resolution CNN, the probability to select a patch on a lower resolution for training is set to be $60\%$ 
of the probability for the respective next higher resolution. 
For lower resolutions, we also use the distribution $N(p^+_2)$. This leads to a more wide spread distribution with respect to the full image resolution.

Feature maps created by our CNNs are not used directly. Instead, we perform a 2x low-pass filter on them, before using them.
Low-pass filtering image data creates matching invariance while increasing ambiguity (by removing high frequency information).  
Assuming that CNNs are unable to create perfect matching invariance, we can expect a similar effect on feature maps created by CNNs.
In fact, a small low-pass filter clearly increases the matching robustness.

The \textit{Flow Fields} approach~\cite{bailer2015flow} uses a secondary consistency check with different patch size. 
With our approach, this would require to train and execute two additional CNNs.
To keep it simple, we perform the secondary check with the same features. This is possible due to the fact that \textit{Flow Fields} is a randomized approach. 
Still, our tests with the original features show that a real secondary consistency check performs better. 
The reasoning for our design decisions in Figure~\ref{ffscales} can be found in Section~\ref{decisions}.

\subsection{Evaluation Methodology for Patch Matching}

In previous works, the evaluation of the matching robustness of (learning based) features was performed by evaluation methods commonly used in classification problems
like ROC in~\cite{brown2011discriminative,zagoruyko2015learning} or PR in~\cite{simo2015discriminative}.
However, patch matching is not a classification problem, but a binary decision problem. 
While one can freely label data in classification problems, patch matching requires 
to choose, at each iteration, out of two proposal patches $p_2, p^*_2$ the one that fits better to $p_1$.
The only exception from this rule is outlier filtering. This is not really an issue, as there are better approaches for outlier filtering, like
the forward backward consistency check~\cite{bailer2015flow}, which is more robust than matching-error based outlier 
filtering\footnote{Even if outlier filtering would be performed by matching error, the actual matching remains a decision problem.}. 
In our evaluation, the \textit{matching robustness} $r$ of a network is determined as the probability that a wrong patch $p^-_2$ is not confused with the correct
 patch $p^+_2$:
\begin{equation}
r = \sum_{ (I_1,I_2) \in S} \sum_{p_1 \in I_1}  P^{r}_{p_1}( p^-_2 )/ (|I_1| |S|)  
\end{equation}
\begin{equation}
\begin{split}
  P^{r}_{p_1}( p^-_2 ) =  P( L_2(p_1,p_2^+)  < L_2(p_1,p^-_2) ),\\
  p_1 \equiv p^+_2 \in I_2,  p_1 \neq p^-_2 \in N(p^+_2), ~~~~~~~~~~
  \end{split}
\end{equation}
where $S$ is a set of considered image pairs $(I_1,I_2)$, $|S|$ the number of image pairs and $|I_1|$ the number of pixels in $I_1$.
As $r$ is a single value we can plot it for different cases: 
\begin{enumerate}\itemsep0.5pt
\item
The curve for different spatial distances between $p^+_2$ and $p^-_2$ ($r_{dist}$).
\item
The curve for different optical flow displacements between $p_1$ and $p^+_2$ ($r_{flow}$). 
\end{enumerate}

$r_{dist}$ and $r_{flow}$ vary strongly for different locations. This makes differences between different networks hard to visualize.
For better visualization, we plot the \textit{relative matching robustness errors} $E_{dist}$ and $E_{flow}$, computed with respect to a pre-selected network $net1$. $E$ is defined as:
\begin{equation}
 E(net1,net2) = (1-r(net2))/(1-r(net1))
\end{equation}

   \begin{table*}[t]
\small
 \centering
 \begin{tabular}{|c|C{1.1 cm}|C{1.1 cm}|C{1.0 cm}|C{1.0 cm}|C{1.0 cm}|C{1.0 cm}|}
   \hline
  Approach  & EPE $>$3 px noc.   & EPE $>$3 px all  & EPE noc. & EPE all    \\
  \hline
  ours &\textbf{4.95\%} & \textbf{11.89\%} & \textbf{1.10 px} & \textbf{2.60 px} \\
  \hline
  original (\cite{bailer2015flow}+CNN)& 5.48\% & 12.59\% & 1.28 px & 3.08 px  \\
  \hline
  \hline
  ms res 1 & 5.17\%  & 12.10\% & 1.17 px & 2.80 px \\
  \hline
  \end{tabular}
   \begin{tabular}{|c|C{1.1 cm}|C{1.1 cm}|C{1.0 cm}|C{1.0 cm}|C{1.0 cm}|C{1.0 cm}|}
   \hline
  Approach  &EPE $>$3 px noc.   & EPE $>$3 px all  & EPE noc. & EPE all    \\
  \hline 
  all resolutions & 5.66\% & 13.01\% & 1.27 px & 2.98 px  \\
  \hline
  nolowpass &\uline{5.21\%} & \uline{12.21\%} & \uline{1.19 px} & \uline{2.80 px}  \\
  \hline
  \hline
  ms res 2+ & 5.18\% & 12.12\% & 1.21 px & 2.84 px \\
  \hline
  \end{tabular}
  \vspace{0.1cm}
  \caption{Comparison of CNN based multi-scale feature creation approaches. 
   See text for details. }
  \label{ffmethods}
  \end{table*}

   \begin{figure*}[t]
  \vspace{0.05cm}
\centering
\subfigure[]{ \includegraphics[width=0.47\linewidth]{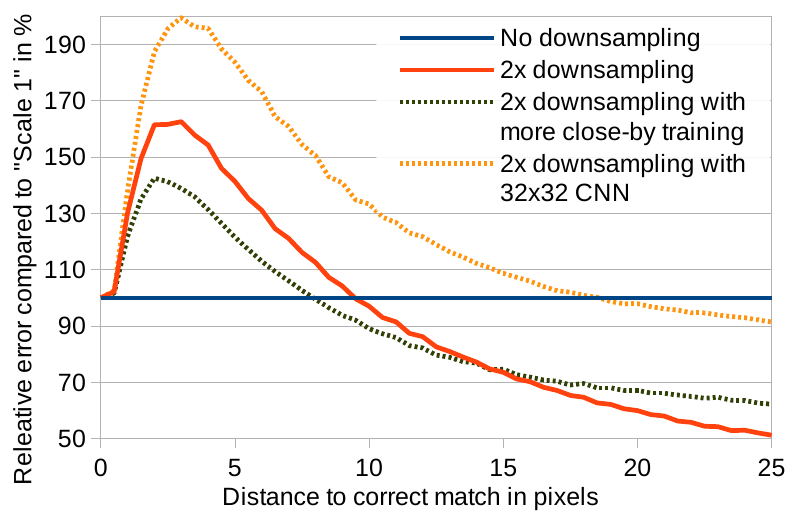}}
\subfigure[]{ \includegraphics[width=0.47\linewidth]{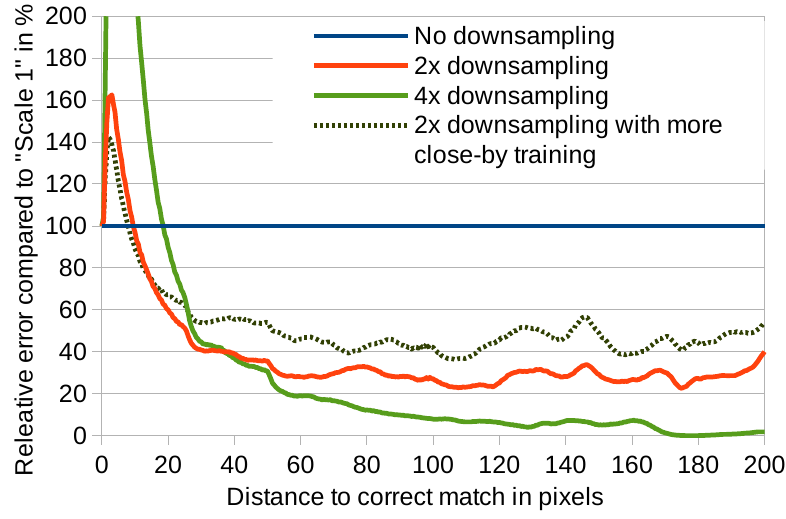}}
\vspace{-0.06cm}
   \caption{Relative matching robustness errors $E_{dist}($``No Downsampling''$,X)$.
   Features created on lower resolutions are more accurate for large distances but less accurate for small ones. \textit{No downsampling}
   is on the horizontal line as results are normalized for it. Details in text.}
\label{scales}
\end{figure*}

\vspace{-0.2cm} 
\section{Evaluation} \label{eval}
We examine our approach on the KITTI 2012 training set~\cite{geiger2013vision} as it is one of the few datasets that contains ground truth for non-synthetic large
displacement optical flow estimation.  
We use patches taken from 130 of the 194 images of the set for training and patches from the remaining 64 images for validation. 
Each tested network is trained with 10 million negative and 10 million positive samples in total.
Furthermore, we publicly validate the performance of our approach by submitting our results to the KITTI 2012, the recently published KITTI 2015~\cite{menze2015object}
and MPI-Sintel evaluation portals (with networks trained on the respective training set).
We use the original parameters of the \textit{Flow Fields} approach~\cite{bailer2015flow} except for the outlier filter distance $\epsilon$ and the random search distance $R$.
$\epsilon$ is set to the best value for each network (with accuracy $\pm 0.25$, mostly: $\epsilon = 1.5$).  
The random search distance $R$ is set to 2 for four iterations and to $R=1$ for two additional iterations to increase accuracy. 
The batch size is set to 100 and $m$ to 1.

To evaluate the quality of our optical flow results we calculate the endpoint error (EPE) for non-occluded areas (noc) as well as occluded + non-occluded areas (all).
(noc) is a more direct measure as CNNs are only trained here. However, the interpolation into occluded areas (like \textit{Flow Fields} we use EpicFlow~\cite{revaud2015epicflow} for that)
 also depends on good matches close to the occlusion boundary, where matching is especially difficult due to partial occlusions of patches.
Furthermore, like~\cite{geiger2013vision}, we measure the percentage of pixels with an EPE above a threshold in pixels (px).

\subsection{Comparison of CNN based Multi-Scale Feature Map Approaches}\label{decisions}
In Table~\ref{ffmethods},  we compare the \textit{original} feature creation approach (Figure~\ref{ffscales} left) with \textit{our} approach (Figure~\ref{ffscales} right), 
with respect to our CNN features.
We also examine two variants of our approach in the table: \textit{nolowpass} which does not contain the ``Low-Pass 2x'' blocks and
\textit{all resolutions} which uses 1x,2x,4x,8x up/downsampling for the four scales (instead of 1x,1x,2x,4x in Figure~\ref{ffscales} right). The reason why \textit{all resolutions} does not work well is 
 demonstrated in Figure~\ref{scales} (a). Starting from a distance between $p^+_2$ and $p^-_2$ of 9 pixels, CNN based features created on a 2x down-sampled image match more robustly 
 than CNN based features created on the full image resolution.
  This is insufficient as the random search distance on scale 2 is only $2R = 4$ pixels. 
  Thus, we use it for scale 3 (with random search distance $4R = 8 \approx 9$ pixels).
 
 One can argue that by training the CNN with more close-by samples $N_{close}(p^+_2)$ more accuracy could be gained.
 But raising extremely the amount of close-by samples only reduces the accuracy threshold from 9 to 8 pixels.
 Using a CNN with smaller 32x32 patches instead of 56x56 patches does not raise the accuracy either-- it even clearly decreases it. 
 Figure~\ref{scales} (b) shows that downsampling decreases the matching robustness error significantly for larger distances. In fact, for a distance above
  170 pixels, the relative error of \textit{4x downsampling} is reduced by nearly 100\% compared to \textit{No downsampling} -- which is remarkable.   

  \renewcommand{\tabcolsep}{1.35pt}
  
  \begin{table}[t]
\small
 \centering
 \begin{tabular}{|C{1.95 cm}|C{1.1 cm}|C{1.2 cm}|C{1.1 cm}|C{1.1 cm}|C{1.2 cm}|C{1.2 cm}|C{1.2 cm}|C{1.2 cm}|}
  \hline
  Approach/ Loss  &EPE $>$3 px noc.   & EPE $>$3 px all  & EPE noc. & EPE all   & robust- ness $r$  \\
  \hline
  $L_h$ &   7.26\% & 14.78\% & 1.46 px & 3.33 px & 98.63\% \\ 
  \hline
  $L_t$, $t=0.2$ &6.17\% & 13.51\% & 1.37 px & 3.10 px & 99.15\%  \\
  \hline 
  $L_t$, $t=0.3$ &\textbf{4.95\%} & \textbf{11.89\%} & \textbf{1.10 px} & \textbf{2.60 px} & 99.34\% \\
  \hline
  $L_t$, $t=0.4$ &\uline{5.18\%} & \uline{12.10\%} & 1.25 px & 3.14 px & \textbf{99.41\%} \\ 
  \hline
  $L_g$, $g=0.2$ & 5.92\% & 13.17\% & 1.41 px & 3.37 px & 99.15\% \\ 
  \hline
  $L_g$, $g=0.4$ & 5.89\% & 13.23\% & 1.41 px & 3.36 px & 99.31\% \\
  \hline 
  $L_g$, $g=0.6$ & 6.37\% & 13.74\% & 1.51 px & 3.40 px & 99.08\% \\ 
  \hline
  Hard Mining x2~\cite{simo2015discriminative} & 6.03\% & 13.34\% & 1.35 px & 2.99 px & 99.07\% \\
  \hline
   DrLIM~\cite{hadsell2006dimensionality} & 5.36 \% & 12.40\%  & \uline{1.18 px} & \uline{2.79 px} & 99.15\% \\
   \hline
   CENT.~\cite{gadot2015patchbatch} & 6.32\% & 13.90\% & 1.46 px & 3.37 px & 98.72\% \\
  \hline 
  \hline 
  SIFTFlow~\cite{liu2008sift}  & 11.52\% & 19.79\% & 1.99 px & 4.33 px  & 97.31\% \\ 
  \hline
  SIFTFlow*\cite{liu2008sift} & 5.85\% & 12.90\% & 1.52 px & 3.56 px  & 97.31\% \\ 
  \hline
 \end{tabular}

 \vspace{0.1cm}
 \caption{Results on KITTI 2012~\cite{geiger2013vision} validation set. 
 Best result is bold, 2. best underlined.  \textit{SIFTFlow} uses our pipeline tailored for CNNs. \textit{SIFTFlow*} uses the original pipeline~\cite{bailer2015flow} (Figure~\ref{ffscales} left).
 }
 \label{trainkitti}
\end{table}

\paragraph{Multi-resolution network training} We examine three variants of training our multi-resolution network (green boxes in Figure~\ref{ffscales}): training it on $100\%$, $50\%$ and $25\%$ resolution 
although it is only used for $50\%$ and $25\%$ resolution, at testing time (\textit{ours} in Table~\ref{ffmethods}), 
training it on $50\%$ and $25\%$ resolutions, where it is used for at testing time (\textit{ms res 2+}) 
and training it only on $100\%$ resolution (\textit{ms res 1}). 
As can be seen in Table~\ref{ffmethods} training on all resolutions (\textit{ours}) clearly performs best. 
Likely, mixed training data performs best as samples of the highest resolution provide the largest entropy while samples of lower resolutions fit better to the problem.
However, training samples of lower resolutions seem to harm training for higher resolutions. 
Therefore, we use an extra CNN for the highest resolution.

\subsection{Loss Functions and Mining}\label{s42}
\begin{figure*}[t]
\centering
\subfigure[by distance between $p^+_2$ and $p^-_2$]{ \includegraphics[width=0.47\linewidth]{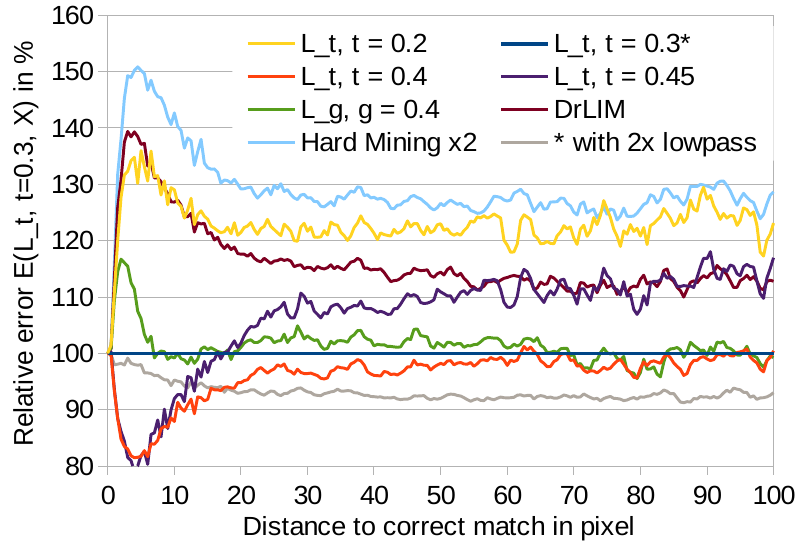}}
\subfigure[by flow displacement (offset between $p_1$ and $p^+_2$)]{ \includegraphics[width=0.47\linewidth]{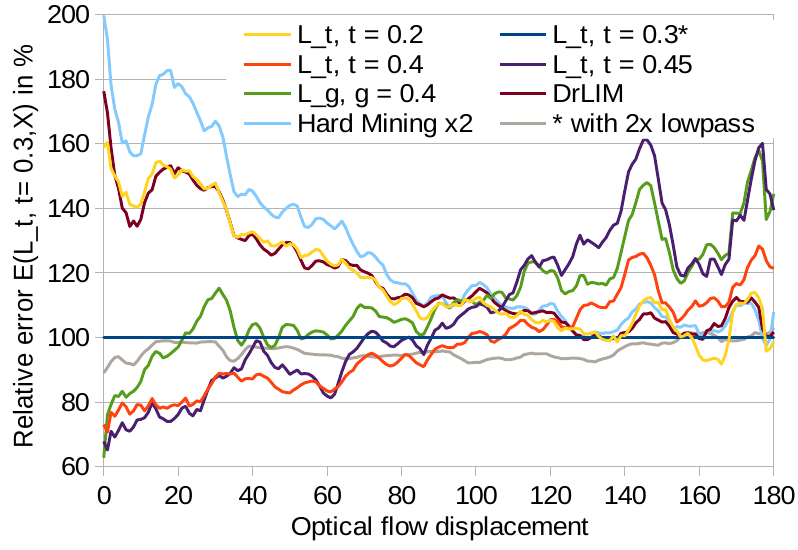}}
   \caption{Relative matching robustness errors $E(``L_t, t=0.3'',X)$ for different loss functions plotted for different distances (a) and displacements (b).
     Note that the plot for $L_t, t=0.3$ is on the horizontal line, as $E$ is normalized for it. See text for details. }
\label{losses} 
\end{figure*}

We compare our loss $l_t$ to  other state-of-the-art losses and Hard Mining~\cite{simo2015discriminative} in Figure~\ref{losses} and Table~\ref{trainkitti}.
As shown in the table, our thresholded loss $l_t$ with $t=0.3$ clearly outperforms all other losses.
DrLIM~\cite{hadsell2006dimensionality} reduces the mentioned flaw in the hinge loss, by training samples with small hinge loss less.   
While this clearly reduces the error compared to hinge, it cannot compete with our thresholded loss $l_t$.
Furthermore, no speedup during training is possible like with our approach. 
CENT. (CENTRIFUGE)~\cite{gadot2015patchbatch} is a variant of DrLIM which performs worse than DrLIM in our tests. 

Hard Mining~\cite{simo2015discriminative} only trains the hardest samples with the largest hinge loss and thus also speeds up training.
However, the percentage of samples trained in each batch is fixed and does not adapt to the requirements of the training data like in our approach. 
With our data, Hard Mining becomes unstable with a mining factor above 2 i.e. the loss of negative samples becomes much larger than the loss
of positive samples. This leads to poor performance ($r = 96.61\%$ for Hard Mining x4). We think this has to do with the fact that the hardest
of our negative samples are much harder to train than the hardest positive samples. 
Some patches are e.g. fully white due to overexposure (negative training has no effect here).
Also, many of our negative samples have, in contrast to the samples of~\cite{simo2015discriminative}, a very small spatial distance to their positive counterpart.  
This makes their training even harder (We report most failures for small distances, see supplementary material), while positive samples do not change.

     \begin{figure}[t]
\centering
 \includegraphics[width=1.00\linewidth]{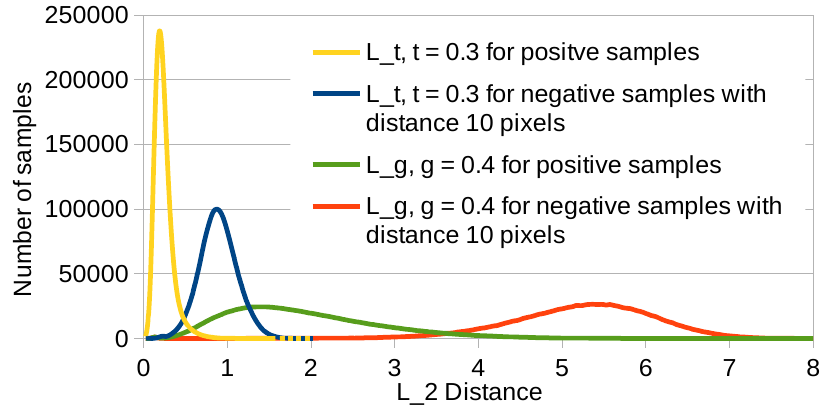}
   \caption{ The distribution of $L_2$ errors for different for $L_t$ and $L_g$ for positive samples $p^+_2$ and negative samples $p^-_2$ with distance of 10 pixels to the
corresponding positive sample.}
\label{variance}
\end{figure}

To make sure that our dynamic loss based mining approach ($L_t$ with $t =0.3$) cannot become unstable towards much larger negative loss values we tested it to an extreme: 
we randomly removed 80\% of the negative training samples while keeping all positive. 
Doing so, it not only stayed stable, but it even used a smaller positive/negative sample mining ratio than the approach with all training samples -- possibly 
it can choose harder positive samples which contribute more to training. Even with the removal of 80\% (8 million) of possible samples we achieved
a matching robustness $r$ of $99.18\%$.

$L_g$ performed best for $g=0.4$ which corresponds to a gap of $L_t, t=0.3$ ($g_{L_t} = 1-2t$).
However, even with the best $g$, $L_g$ performs significantly worse than $L_t$.
This is probably due to the fact that the variance $\mathrm{Var} (L_2(p_1,p_2))$ is much larger for $L_g$ than for $L_t$.
As shown in Figure~\ref{variance}, this is the case for both positive ($p^+_2$) as well as negative ($p^-_2$) samples.
We think this affects the test set negatively as follows: if we assume that $p_1$, $p^+_2$, $p^-_2$ are unlearned test set patches it is clear that the condition $L_2(p_1,p^+_2) < L_2(p_1,p^-_2)$ is more likely violated 
if $\mathrm{Var} (L_2(p_1,p^+_2))$ and $\mathrm{Var} (L_2(p_1,p^-_2))$ are large compared to the learned gap. 
Only with $L_t$ it is possible to force the network to keep the variance small compared to the gap. 
With $L_g$ it is only possible to control the gap but not the variance, while $l_h$ keeps the variance small but cannot limit the gap.

\paragraph{Matching Robustness plots}Some loss functions perform worse than others although they have a larger matching robustness $r$.
This mostly can be explained by the fact that 
they perform poorly for large displacements (as shown in Figure~\ref{losses} (b)).
Here, correct matches are usually more important as missing matches lead lo larger endpoint errors.
An averaged $r$ over all pixels does not consider this.

Figure~\ref{losses} also shows the effect of parameter $t$ in $L_t$. Up to $ t\approx0.3$, all distances and flow displacements are improved,
while small distances and displacements benefit more and up to a larger $t\approx0.4$. 
The improvement happens as unnecessary destructive training is avoided (see Section~\ref{s31}).
Patches with small distances benefit more form larger $t$, likely as the \textit{real gap} 
$ g_{real} = |L_2(p_1,p^-_2) - L_2(p_1,p^+_2)|$ is smaller here (as $p^-_2$ and  $p^+_2$ are very similar for small distances).
For large displacements patches get more chaotic (due to more motion blur, occlusions etc.), which forces larger variances of the
$L_2$ distances and thus a larger gap is required to counter the larger variance.

$L_g$ performs worse than $L_t$ mainly at small distances and large displacements. 
Likely, the larger variance is more destructive for small distances, as the \textit{real gap} $g_{real}$ is smaller (more sensitive) here.
Figure~\ref{losses} also shows that low-pass filtering the feature map increases the matching robustness for all distances and displacements.
In our tests, a $2.25 \times$ low-pass performed the best (tested with $\pm 0.25$).
Engineered SIFTFlow features can benefit from much larger low-pass filters which makes the original pipeline (Figure~\ref{ffscales} left) extremely efficient for them.
However, using them with our pipeline (which recalculates features on different resolutions) shows that their low matching robustness is justified (see Table~\ref{trainkitti}).
SIFTFlow also performs better in outlier filtering. 
Due to such effects that can so far not directly be trained, it is still challenging to beat well designed purely heuristic approaches with learning. 
In fact, existing CNN based approaches often still underperform purely heuristic approaches -- even direct predecessors (see Section~\ref{s43}).

\renewcommand{\tabcolsep}{3.0pt}

\begin{table*}[t]
\small
 \centering
  \begin{tabular}{|c|C{2.1 cm}|C{2.1 cm}|C{2.1 cm}|C{2.1 cm}|C{1.3 cm}|C{1.3 cm}|C{1.3 cm}|}
  \hline
  Method  &EPE $>$3 px noc.&   EPE $>$5 px noc.& EPE $>$3 px all &EPE $>$5 px all & EPE noc. & EPE all & runtime \\
  \hline 
  Ours (56x56) & \textbf{4.89 \%} & \textbf{3.04 \%} & \textbf{13.01 \%} & \textbf{9.06 \%} & \textbf{1.2 px} & \textbf{3.0 px} & \textbf{23s} \\ 
  \hline 
  PatchBatch (71x71) \cite{gadot2015patchbatch}& \uline{4.92} \% &  \uline{3.31} \% &  \uline{13.40} \% &  \uline{10.18} \% & \textbf{ 1.2 px} &  \uline{3.3 px} & 60s \\ 
  \hline
  PatchBatch (51x51) \cite{gadot2015patchbatch}& 5.29 \% & 3.52 \% & 14.17 \% & 10.36 \% & 1.3 px & 3.3 px & \uline{50s} \\ 
  \hline
  Flow Fields \cite{bailer2015flow} &  5.77 \% & 3.95 \% & 14.01 \% & 10.21\% &  1.4 px & 3.5 px  & \textbf{23s}\\
   \hline
  PatchBatch*(51x51)\cite{gadot2015patchbatch} & 5.94\% \cite{gadot2015patchbatch} & - & -&-&-&-  & 25.5s \cite{gadot2015patchbatch} \\
  \hline 
  \end{tabular}
  \vspace{0.1cm}
   \caption{Results on KITTI 2012~\cite{geiger2013vision} test set. Numbers in brackets show the patch size for learning based methods. Best result for published methods
 is bold, 2. best is underlined. PatchBatch* is PatchBatch without pixel-wise batch normalization.  }
  \vspace{0.3cm}
   \label{trkit12}
 \renewcommand{\tabcolsep}{5.0pt}
 \begin{tabular}{|C{2.0 cm}|c|C{1.4 cm}|C{1.2 cm}|C{1.2 cm}|C{1.2 cm}|C{1.2 cm}|C{1.2 cm}|C{1.2 cm}|C{1.2 cm}|C{1.2 cm}|}
  \hline
  & & \multicolumn{2}{c|}{background} & \multicolumn{2}{c|}{foreground (cars)} & \multicolumn{2}{c|}{total} &\\
  \hline
 Type & Method  &EPE $>$3 px noc.&   EPE $>$3 px all  &EPE $>$3 px noc.&   EPE $>$3 px all  &EPE $>$3 px noc.&   EPE $>$3 px all   & runtime \\
  \hline
 \multirow{ 3}{*}{\begin{minipage}{0.8in}\centering Rigid Segmentation based methods\end{minipage}} & SDF \cite{bai2016exploiting}     &   \textbf{5.75\%}  & \textbf{8.61\%}  & 22.28\% & 26.69\% & \textbf{8.75\%} & \textbf{11.62\%} & unknown\\
  \cline{2-9}
   &  JFS \cite{hur2016joint}	&  \textbf{7.85\%}& \textbf{15.90\%} &	\textbf{18.66\%} 	& \textbf{22.92\%}  &	\textbf{9.81\%	}&  \textbf{17.07\%} & 13 min \\
    \cline{2-9}
 & SOF~\cite{sevilla2016optical} & \textbf{8.11\%} &  \textbf{14.63\%} & 23.28\% & 27.73\%	& \textbf{10.86\%} &   \textbf{16.81\% } & 6 min \\ 
     \hline
    \hline
\multirow{ 4}{*}{\begin{minipage}{0.8in}\centering General methods\end{minipage}}  &  Ours (56x56) & \textbf{8.91\%} &  \textbf{18.33\%} &   \textbf{20.78\%} & \textbf{24.96\%} &  \textbf{11.06\%} & \textbf{19.44\%} & \textbf{23s} \\ 
   \cline{2-9}
&  PatchBatch (51x51) \cite{gadot2015patchbatch}&  10.06\%  & \uline{19.98\%}  &	26.21\% &30.24\%  &	12.99\% & \uline{21.69\%}   & \uline{50s} \\ 		 	
 \cline{2-9}
&  DiscreteFlow~\cite{menze2015discrete} & \uline{9.96\%} & 21.53 \%  & \uline{22.17\%} & \uline{26.68 \%} & \uline{12.18\%} & 22.38\% & 3 min \\
 \cline{2-9}
&  DeepDiscreteFlow~\cite{Guney2016ACCV} & 10.44\% & 20.36 \%  & 25.86\% & 29.69 \% & 13.23\% & 21.92\% & 1 min \\

  \hline
 \end{tabular}
 \vspace{0.1cm}
 \caption{Results on KITTI 2015~\cite{menze2015object} test set. Numbers in brackets shows the used patch size for learning based methods. 
  Best result for all published general optical flow methods is bold, 2. best underlined.  
  Bold for segmentation based method shows that the result is better than the best general method.
  Rigid segmentation based methods were designed for urban street scenes and similar 
  containing only segmentable rigid objects and rigid background (and are usually very slow), while general methods work for all optical flow problems. 
  }
 \label{trkit15}
\end{table*}

\subsection{Public Results}\label{s43}
Our public results on the KITTI 2012~\cite{geiger2013vision}, 2015~\cite{menze2015object} and MPI-Sintel~\cite{butler2012naturalistic} 
evaluation portals are shown in Table~\ref{trkit12},~\ref{trkit15} and~\ref{trsint}. 
For the public results we used 4 extra iterations with $R=1$ for best possible subpixel accuracy and for similar runtime to \textit{Flow Fields}~\cite{bailer2015flow}.
$t$ is set to 0.3. 
On KITTI 2012 our approach is the best in all measures, although we use a smaller 
patch size than \textit{PatchBatch (71x71)}~\cite{gadot2015patchbatch}. \textit{PatchBatch (51x51)} with a patch size more similar to ours performs even worse.
\textit{PatchBatch*(51x51)} which is like our work without pixel-wise batch normalization even trails purely heuristic methods like \textit{Flow Fields}.

On KITTI 2015 our approach also clearly outperforms PatchBatch and all other general optical flow methods including DeepDiscreteFlow~\cite{Guney2016ACCV} that, despite using CNNs,
trails its engineered predecessor DiscreteFlow~\cite{menze2015discrete} in many measures.
The only methods that outperform our approach are the rigid segmentation based methods SDF~\cite{bai2016exploiting}, JFS~\cite{hur2016joint} and SOF~\cite{sevilla2016optical}.
These require segmentable rigid objects moving in front of rigid background and are thus not suited for scenes that contain non-rigid 
objects (like MPI-Sintel) or objects which are not easily segmentable.
Despite not making any such assumptions our approach 
outperforms two of them in the challenging foreground (moving cars with reflections, deformations etc.).
Furthermore, our approach is clearly the fastest of all top performing methods although there is still optimization potential (see below). 
Especially, the segmentation based methods are very slow.

On the non rigid MPI-Sintel datasets our approach is the best in the non-occluded areas, which can be matched by our features. 
Interpolation into occluded areas with EpicFlow~\cite{revaud2015epicflow} works less well, which is no surprise as 
 aspects like good outlier filtering which are important for occluded areas are not learned by our approach. 
Still, we obtained the best overall result on the more challenging final set that contains motion blur. 
In contrast, PatchBatch lags far behind on MPI-Sintel, while DeepDiscreteFlow again clearly trails its predecessor DiscreteFlow on the clean set, but not the final set.
Our approach never trails on the relevant matchable (non-occluded) part. 

Our detailed runtime is 4.5s for CNNs (GPU) + 16.5s patch matching (CPU) + 2s for up/downsampling
and low-pass (CPU). The CPU parts of our approach likely can be significantly sped up using GPU versions 
like a GPU based propagation scheme~\cite{bailer2012scale,galliani2015massively} for patch matching.
This is contrary to PatchBatch where the GPU based CNN already takes the majority of time (due to pixel-wise normalization). 
Also, in final tests (after submitting to evaluation portals) we were able to improve our CNN architecture (see supplementary material) so that it only needs 2.5s with
only a marginal change in quality on our validation set.

\section{Conclusion and Future Work}\label{conc}
In this paper, we presented a novel extension to the hinge embedding loss that not only outperforms other losses in learning robust patch representations, but 
also allows to increase the training speed and to be robust with respect to unbalanced training data.
We presented a new multi-scale feature creation approach for CNNs and 
proposed new evaluation measures by plotting matching robustness with respect to patch distance and motion displacement. 
Furthermore, we showed that low-pass filtering feature maps created by CNNs improves the matching result. 
All together, we proved the effectiveness of our approach by submitting it to the KITTI 2012, KITTI 2015 and MPI-Sintel evaluation portals where we, 
as the first learning based approach, achieved state-of-the-art results on all three datasets.
Our results also show the transferability of our contribution, as our findings 
 made in Section~\ref{decisions} and ~\ref{s42} (on which our architecture is based on) are solely based on KITTI 2012 validation set, 
but still work unchanged on KITTI 2015 and MPI-Sintel test sets, as well. 

\renewcommand{\tabcolsep}{0.5pt}

\begin{table}[t]
\small
 \centering
  \begin{tabular}{|c|C{1.4 cm}|C{1.85 cm}|C{1.85 cm}|C{1.8 cm}|C{1.8 cm}|C{1.3 cm}|C{1.3 cm}|}
  \hline
 Method(final) & EPE all & EPE not occl. & EPE occluded\\
 \hline
 Ours &  \textbf{5.363} &  \textbf{2.303} &  \uline{30.313} \\
 \hline
 DeepDiscreteFlow\cite{Guney2016ACCV} & \uline{5.728} & 2.623 & 31.042 \\
 \hline
 FlowFields \cite{bailer2015flow}&	5.810 &	\uline{2.621} &	31.799 \\
 \hline
 CPM-Flow \cite{huefficient}& 	5.960 &	2.990 &	\textbf{30.177} \\
 \hline
 DiscreteFlow~\cite{menze2015discrete} & 6.077 & 2.937 & 31.685\\
 \hline
 PatchBatch \cite{gadot2015patchbatch}&	6.783 &	3.507 &	33.498 \\
 \hline
 \hline
 Method(clean) & EPE all & EPE not occl. & EPE occluded\\
 \hline
 CPM-Flow \cite{huefficient}&	\textbf{3.557} &	1.189 &	\textbf{22.889} \\
 \hline
 DiscreteFlow~\cite{menze2015discrete} &	\uline{3.567} &	1.108 & 23.626 \\
 \hline
 FullFlow~\cite{chen2016full} & 3.601 & 1.296 & 	\uline{22.424} \\
 \hline
 FlowFields \cite{bailer2015flow}&	3.748 &	\uline{1.056} &	25.700 \\
 \hline
 Ours & 	3.778 &	\textbf{0.996} &	26.469 \\
 \hline
 DeepDiscreteFlow\cite{Guney2016ACCV} & 3.863 & 1.296 & 24.820 \\
 \hline
 PatchBatch \cite{gadot2015patchbatch}& 	5.789 &	2.743 &	30.599 \\
  \hline
 \end{tabular}
 \vspace{0.1cm}
 \caption{Results on MPI-Sintel~\cite{butler2012naturalistic}. Best result for all published methods is bold, second best is underlined.  }
 \label{trsint}
\end{table}

In future work, we want to improve our network architecture (Table~\ref{arch}) by using techniques 
like (non pixel-wise) batch normalization and dilated convolutions~\cite{luo2016efficient}.
Furthermore, we want to find out if low-pass filtering invariance also helps in other application, like sliding window object detection~\cite{ren2015faster}.
We want to further improve our loss function $L_t$ e.g. by a dynamic $t$ that depends on the properties of training samples.
So far, we just tested a patch size of 56x56 pixels, although~\cite{gadot2015patchbatch} showed that larger patch sizes can perform even better.
It might be interesting to find out which is the largest beneficial patch size.
Frames of MPI-Sintel with very large optical flow showed to be especially challenging.
They lack training data due to rarity, but still have a large impact on the average EPE (due to huge EPE).
We want to create training data tailored for such frames and examine if learning based approaches benefit from it. 

\subsection*{Acknowledgments}
This work was funded by the BMBF project DYNAMICS (01IW15003).

{\small
\bibliographystyle{ieee}
\bibliography{arxiv_2}
}

\end{document}